\pdfoutput=1

\documentclass[11pt]{article}

\usepackage{ACL2023}

\usepackage{times}
\usepackage{latexsym}
\usepackage{subcaption}

\usepackage[T1]{fontenc}

\usepackage[utf8]{inputenc}

\usepackage{microtype}

\usepackage{inconsolata}
\usepackage{enumitem}
\usepackage{booktabs}
\usepackage{graphicx}
\usepackage{multirow}
\usepackage{csquotes}
\usepackage{amsfonts}
\usepackage{hhline}
\usepackage{arydshln}
\usepackage{amsmath,bm}
\usepackage{colortbl}
\usepackage{twemojis}

\newcommand{\brick}[1]{\textcolor{BrickRed}{#1}}
\newcommand{\olive}[1]{\textcolor{OliveGreen}{#1}}

\title{\textsc{MTCue}: Learning Zero-Shot Control of Extra-Textual Attributes \\by Leveraging Unstructured Context in Neural Machine Translation}

\author{Sebastian Vincent \and Robert Flynn \and Carolina Scarton \\ 
        Department of Computer Science, University of Sheffield, UK\\
        \texttt{\{stvincent1,rjflynn2,c.scarton\}@sheffield.ac.uk}\\
}

\begin{document}

\maketitle
\begin{abstract}
Efficient utilisation of both intra- and extra-textual context remains one of the critical gaps between machine and human translation. Existing research has primarily focused on providing individual, well-defined types of context in translation, such as the surrounding text or discrete external variables like the speaker's gender. This work introduces \textsc{MTCue}, a novel neural machine translation (NMT) framework that interprets all context (including discrete variables) as text. \textsc{MTCue} learns an abstract representation of context, enabling transferability across different data settings and leveraging similar attributes in low-resource scenarios. With a focus on a dialogue domain with access to document and metadata context, we extensively evaluate \textsc{MTCue} in four language pairs in both translation directions. Our framework demonstrates significant improvements in translation quality over a parameter-matched non-contextual baseline, as measured by \textsc{bleu} ($+0.88$) and \textsc{Comet} ($+1.58$). Moreover, \textsc{MTCue} significantly outperforms a \enquote{tagging} baseline at translating English text. Analysis reveals that the context encoder of \textsc{MTCue} learns a representation space that organises context based on specific attributes, such as formality, enabling effective zero-shot control. Pre-training on context embeddings also improves \textsc{MTCue}'s few-shot performance compared to the \enquote{tagging} baseline. Finally, an ablation study conducted on model components and contextual variables further supports the robustness of \textsc{MTCue} for context-based NMT. 

\end{abstract}

\hspace{9mm}
\begin{minipage}[c]{4.5mm}
\includegraphics[width=\linewidth]{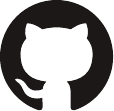}
\end{minipage}
\hspace{0mm}
\begin{minipage}[c]{0.7\textwidth}
\fontsize{8pt}{8pt}\selectfont
\href{https://github.com/st-vincent1/MTCue}
{\texttt{github.com/st-vincent1/MTCue}}
\end{minipage}

\section{Introduction}

Research in neural machine translation (NMT) has advanced considerably in recent years, much owing to the release of the Transformer architecture \citep{Vaswani2017}, subword segmentation \citep{Sennrich2016} and back-translation \citep{Sennrich2016b}. This resulted in claims of human parity in machine translation \citep{Hassan2018},
which in turn prompted researchers to look beyond the sentence level: at how a translation still needs to be compatible with the context it arises in.
\begin{figure}[ht!]
\centering
\resizebox{0.85\linewidth}{!}{
\includegraphics[width=\linewidth]{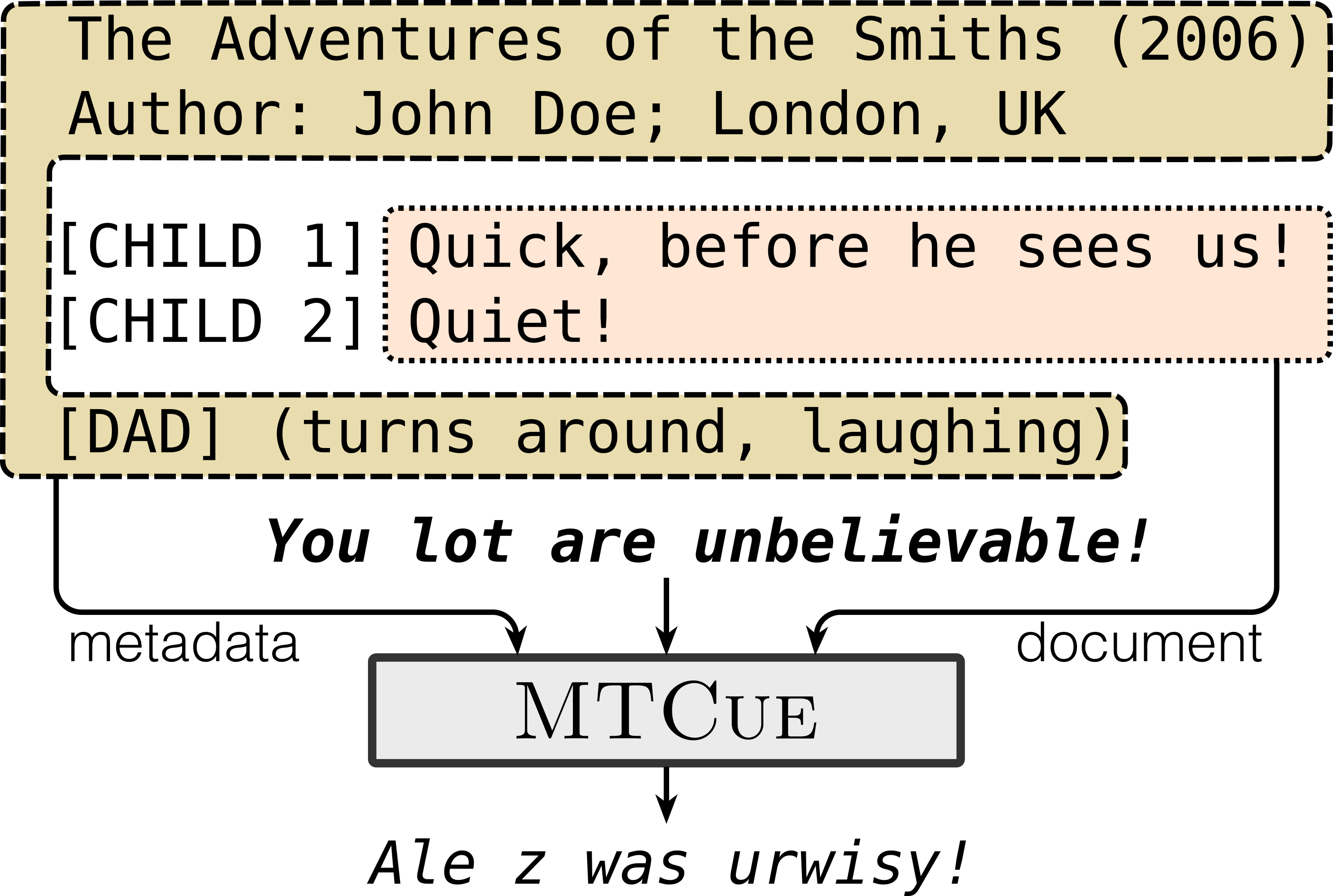}}
\caption{A high-level overview of \mbox{\textsc{MTCue}} (\textsc{en}$\rightarrow$\textsc{pl}).}
\label{fig:system}
\end{figure}

The task of contextual adaptation to more nuanced extra-textual variables like the description of the discourse situation has been largely overlooked, in spite of earlier work suggesting that conversational machine translation may benefit from such fine-grained adaptations \citep{van-der-wees-etal-2016-measuring}. Most existing work on contextual NMT has focused on document-level context instead, aiming to improve the coherence and cohesion of the translated document \citep[e.g.][]{tiedemann-scherrer-2017-neural}. Some research has successfully adapted NMT to extra-textual context variables using supervised learning frameworks on labelled datasets, targeting aspects such as gender \citep{Vanmassenhove2020, Moryossef2019, vincent-etal-2022-controlling-extra}, formality \citep{Sennrich2016a, nadejde-etal-2022-cocoa}, translators' or speakers' style \citep{Michel2018, wang-etal-2021-towards} and translation length \citep{lakew-etal-2019-controlling}, sometimes controlling multiple attributes simultaneously \citep{schioppa-etal-2021-controlling, vincent-etal-2022-controlling-extra}. However, to our knowledge, no prior work has attempted to model the impact of continuous extra-textual contexts in translation or combined the intra- and extra-textual contexts in a robust framework. This is problematic since translating sentences without or with incomplete context is akin to a human translator working with incomplete information. Similarly, only a handful of earlier research has contemplated the idea of controlling these extra-textual attributes in a zero-shot or few-shot fashion \citep{Moryossef2019, anastasopoulos-etal-2022-findings}; such approaches are essential given the difficulty of obtaining the labels required for training fully supervised models.

In some domains, extra-textual context is paramount and NMT systems oblivious to this information are expected to under-perform. For instance, for the dubbing and subtitling domain, where translated shows can span different decades, genres, countries of origin, etc., a one-size-fits-all model is limited by treating all input sentences alike. In this domain, there is an abundance of various metadata (not just document-level data) that could be used to overcome this limitation. However, such adaptation is not trivial: (i) the metadata often comes in quantities too small for training and with missing labels; (ii) it is expressed in various formats and types, being difficult to use in a standard pipeline; (iii) it is difficult to quantify its exact (positive) effect.

In this paper, we address (i) and (ii) by proposing \mbox{\textsc{MTCue}} (\textbf{M}achine \textbf{T}ranslation with \textbf{C}ontextual \textbf{u}niversal \textbf{e}mbeddings), a novel NMT framework that bridges the gap between training on discrete control variables and intra-textual context as well as allows the user to utilise metadata of various lengths in training, easing the need for laborious data editing and manual annotation (\autoref{fig:system}). During inference, when context is provided verbatim, \mbox{\textsc{MTCue}} falls back to a code-controlled translation model; by vectorising the inputs, it exhibits competitive performance for noisy phrases and learns transferrability across contextual tasks.
While (iii) is not directly addressed, our evaluation encompasses two translation quality metrics and two external test sets of attribute control, showing the impact on both translation quality and capturing relevant contextual attributes.

\mbox{\textsc{MTCue}} can generalise to unseen context variables, achieving $100$\% accuracy at a zero-shot formality controlling task; it learns to map embeddings of input contexts to discrete phenomena (e.g. formality), increasing explainability; and it exhibits more robust few-shot performance at multi-attribute control tasks than a \enquote{tagging} baseline. 

The main contributions of this work are:

\begin{enumerate}[noitemsep,nolistsep,leftmargin=*]
    \item \mbox{\textsc{MTCue}} (\S \ref{sec:arch}): a novel framework for \textbf{combining (un)structured intra- and extra-textual context in NMT}  that significantly improves translation quality for four language pairs in both directions: English (\textsc{en}) to/from German (\textsc{de}), French (\textsc{fr}), Polish (\textsc{pl}) and Russian (\textsc{ru}).
    \item A comprehensive evaluation, showing that \mbox{\textsc{MTCue}} can be primed to exhibit \textbf{excellent zero-shot and few-shot performance} at downstream contextual translation tasks (\S \ref{results} and \S \ref{ablation}).
    \item Pre-trained models, code, and an organised version of the OpenSubtitles18 \citep{lison-etal-2018-opensubtitles2018} dataset \textbf{with the annotation of six metadata} are made available.
\end{enumerate}

This paper also presents the experimental settings (\S \ref{exp}), related work (\S \ref{related}) and conclusions (\S \ref{conclusions}). 

\section{Proposed Architecture: \mbox{\textsc{MTCue}}}
\label{sec:arch}

\begin{figure*}[h!]
\centering
\scalebox{0.6}{
\includegraphics[width=\linewidth]{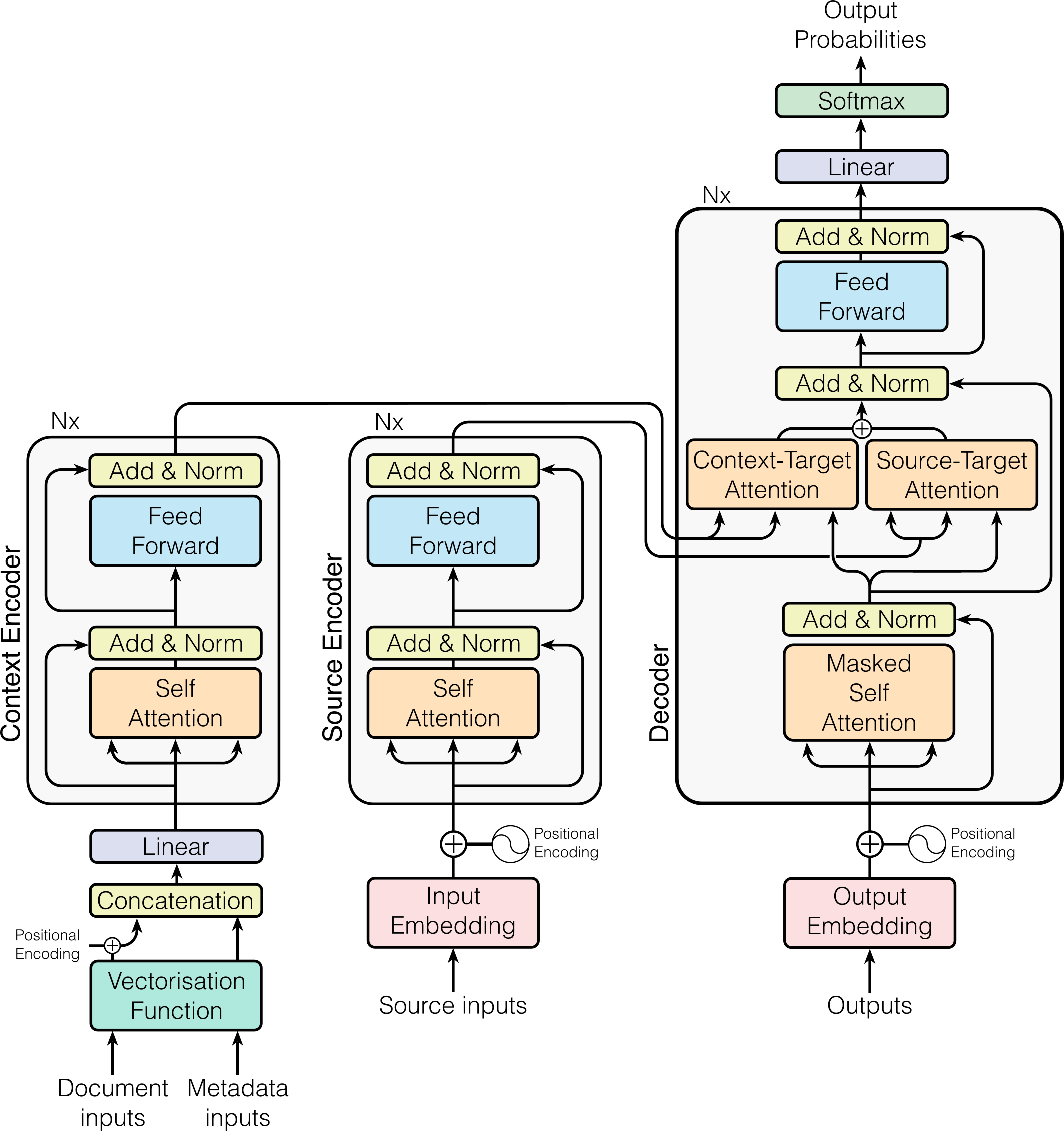}}
\caption{The \mbox{\textsc{MTCue}} architecture. Stylised after the Transformer architecture figure in \citep{Vaswani2017}.}
\label{fig:arch}
\end{figure*}

\textsc{MTCue} is an encoder-decoder Transformer architecture with two encoders: one dedicated for contextual signals and one for inputting the source text. The signals from both encoders are combined using parallel cross-attention in the decoder. Below we describe how context inputs are treated in detail, and later in \S \ref{cxt-enc} and \S \ref{cxt-inc} we describe the context encoder and context incorporation, respectively.

\subsection{Vectorising Contexts} 
\label{sentenceembeddings}
Context comes in various formats: for example, the speaker's gender or the genre of a film are often supplied in corpora as belonging to sets of pre-determined discrete classes, whereas plot descriptions are usually provided as plain text (and could not be treated as discrete without significant loss of information). To leverage discrete variables as well as short and long textual contexts in a unified framework, we define a \textbf{vectorisation function} that maps each context to a single meaningful vector, yielding a matrix $\mathbf{E}_{c \times r}$, where $c$ is the number of contexts and $r$ is the embedding dimension. The function is deterministic (the same input is always embedded in the same way) and semantically coherent (semantically similar inputs receive similar embeddings). We use a sentence embedding model \citep{reimers-2019-sentence-bert} for vectorisation, which produces embeddings both deterministic and semantically coherent. Motivated by \citet{khandelwal-etal-2018-sharp} and \citet{oconnor-andreas-2021-context} who report that generation models mostly use general topical information from past context, ignoring manipulations such as shuffling or removing non-noun words, we hypothesise that sentence embeddings can effectively compress the relevant context information into a set of vectors, which, when processed together within a framework, will formulate an abstract representation of the dialogue context. We select the \textsc{MiniLMv2} sentence embedding model \citep{wang-etal-2021-minilmv2}, which we access via the \texttt{sentence-transformers} library;\footnote{\url{https://sbert.net/}, accessed 1/5/23.} a similar choice was made concurrently in \citet{vincent2023personalised}. In the experiments, we also refer to \textsc{DistilBERT} \citep{Sanh2019} which is used by one of our baselines, and a discrete embedding function which maps unique contexts to the same embeddings but has no built-in similarity feature.

For any sample, given a set of its $k$ textual contexts $C = [c_1,...c_k]$, we vectorise each one separately using the method described above. The resulting array of vectors is the input we supply to the context encoder in \textsc{MTCue}.

\subsection{Context Encoder}
\label{cxt-enc}
\paragraph{Processing vectorised contexts} The context encoder of \mbox{\textsc{MTCue}} is a standard self-attention encoder with a custom input initialisation. Its inputs are sentence embeddings of context (\S \ref{sentenceembeddings}) projected to the model's dimensions with a linear layer ($384 \rightarrow d_{model}$). In preliminary experiments, we observe that the first layer of the context encoder receives abnormally large input values, which sometimes leads to the explosion of the query ($\mathbf{Q}$) and key ($\mathbf{K}$) dot product $\mathbf{QK}^\text{T}$. We prevent this by replacing the scaled dot product attention with query-key normalisation \citep{henry-etal-2020-query}: applying L2 normalisation of $\mathbf{Q}$ and $\mathbf{K}$ before the dot product, and replacing the scaling parameter $\sqrt{d}$ with a learned one, initialised to a value based on training data lengths.\footnote{An alternative solution applies layer normalisation to the input of the first layer, but we found that this degraded performance w.r.t. \textsc{QK-Norm}.}

\paragraph{Positional embeddings} We use positional context embeddings to (a) indicate the distance of a past utterance to the source sentence and (b) to distinguish metadata inputs from document information. In particular, when translating the source sentence $s_i$ at position $i$ in the document, a sentence distance positional embedding ($POS$) is added to the embedding representations of each past sentence $s_{i-j}$, with $j \in [0,t]$ where $t$ is the maximum allowed context distance:
$e'(s_{i-j}, j) = e(s_{i-j}) + POS(j)$. 
Metadata contexts (${m_0, \ldots, m_n}$) do not receive positional embeddings since their order is irrelevant. The final vectorised input of the context encoder is:
\noindent \scalebox{0.9}{$e'(s_{i},0), e'(s_{i-1}, 1), \ldots, e'(s_{i-t}, t), e(m_0), \ldots, e(m_n).$}

\subsection{Context incorporation}
\label{cxt-inc}

The outputs of the context and source encoders (respectively $\mathcal{C}$ and $\mathcal{S}$) are combined in the decoder using \textbf{parallel attention} \citep{libovicky-etal-2018-input}. Let the output of the decoder self-attention be $\mathcal{T}$. Let $\mathcal{T}_{out} = \text{FFN}(\mathcal{T'}) + \mathcal{T'}$, where $\mathcal{T'}$ is the multi-head attention output; i.e. $\mathcal{T}_{out}$ is $\mathcal{T'}$ with the feed-forward layer and the residual connection applied. In a non-contextual Transformer, source and target representations are combined with cross-attention:
\begin{equation*}
\begin{split}
\mathcal{T'} &= \text{mAttn}(kv = \mathcal{S}, q=\mathcal{T})
\end{split}
\end{equation*}
In contrast, parallel attention computes individual cross-attention of $\mathcal{T}$ with $\mathcal{S}$ and $\mathcal{C}$ and then adds them together:
\begin{equation*} \label{parallel}
\begin{split}
\mathcal{S'} &= \text{mAttn}(kv=\mathcal{S}, q=\mathcal{T}) \\
\mathcal{C'} &= \text{mAttn}(kv = \mathcal{C}, q=\mathcal{T}) \\
\mathcal{T}' &= \mathcal{C'} + \mathcal{S'}
\end{split}
\end{equation*}
Parallel attention is only one of many combination strategies which can be used, and in preliminary experiments we found the choice of the strategy to have a minor impact on performance.

\section{Experimental Setup} 
\label{exp}
\subsection{Data: the OpenSubtitles18 Corpus}

\begin{table}[ht]
\centering
\scalebox{0.7}{
\begin{tabular}{rcccc}
\toprule
Data type & \textsc{en$\leftrightarrow$de} & \textsc{en$\leftrightarrow$fr} & \textsc{en$\leftrightarrow$pl} & \textsc{en$\leftrightarrow$ru} \\
\midrule
\hspace{.5em}Source \& target & $5.3$M & $14.7$M & $12.9$M & $12.4$M \\
\midrule
\textit{\textbf{metadata}} &  &  &  & \\
\hspace{.5em}Genre & $45.3\%$ & $57.8\%$ & $60.5\%$ & $73.4\%$ \\
\hspace{.5em}PG rating & $35.9\%$ & $46.9\%$ & $48.8\%$ & $62.3\%$ \\
\hspace{.5em}Writer(s) & $45.3\%$ & $57.1\%$ & $58.9\%$ & $71.7\%$ \\
\hspace{.5em}Year  & $45.3\%$ & $57.8\%$ & $60.5\%$ & $73.7\%$ \\
\hspace{.5em}Country  & $37.7\%$ & $42.9\%$ & $45.7\%$ & $42.7\%$ \\
\hspace{.5em}Plot description & $43.4\%$ & $57.1\%$ & $59.7\%$ & $72.6\%$ \\
\textit{\textbf{previous dialogue}} &  &  &  & \\
\hspace{.5em} $n-1$ & $60.1\%$ & $68.0\%$ & $63.7\%$  & $73.6\%$ \\
\hspace{.5em} $n-2$ & $42.0\%$ & $51.2\%$ & $46.4\%$ & $57.9\%$ \\
\hspace{.5em} $n-3$ & $31.2\%$ & $40.1\%$ & $35.5\%$ & $46.9\%$ \\
\hspace{.5em} $n-4$ & $23.9\%$ & $32.2\%$ & $28.0\%$ & $38.6\%$ \\
\hspace{.5em} $n-5$ & $18.7\%$ & $26.2\%$ & $22.4\%$ & $32.2\%$ \\
\bottomrule
\end{tabular}
}

\caption{Data quantities for the extracted OpenSubtitles18 corpus. An average of $81$\% samples has at least one context input.}
\label{tab:data-quantity}
\end{table}

\begin{table}[h]
\centering
\scalebox{0.7}{
\begin{tabular}{{p{0.3\linewidth} | p{\linewidth}}}
\toprule
Key & Value \\ \midrule
Source (\textsc{en}) &  This is the Angel of Death, big daddy reaper. \\
Target (\textsc{pl}) & To anioł śmierci. Kosiarz przez wielkie "k". \\
PG rating & PG rating: TV-14 \\
Released & Released in 2009 \\
Writers & Writers: Eric Kripke, Ben Edlund, Julie Siege \\
Plot & Dean and Sam get to know the whereabouts of Lucifer and want to hunt him down. But Lucifer is well prepared and is working his own plans. \\
Genre & Drama, Fantasy, Horror \\
Country & United States, Canada \\ \bottomrule
\end{tabular}}
\caption{Example of a source-target pair and metadata in \textsc{OpenSubtitles}.}
\label{tab:data-sample}
\end{table}

The publicly available OpenSubtitles18\footnote{Created from data from \url{https://opensubtitles.org}.} corpus \citep{lison-etal-2018-opensubtitles2018}, hereinafter \textsc{OpenSubtitles}, is a subtitle dataset in \texttt{.xml} format with IMDb ID attribution and timestamps. It is a mix of original and user-submitted subtitles for movies and TV content. Focusing on four language pairs (\textsc{en}$\leftrightarrow$\{\textsc{de,fr,pl,ru}\}), we extract parallel sentence-level data with source and target document-level features (up to 5 previous sentences) using the timestamps (see \autoref{app:data-prep}). We also extract a range of metadata by matching the IMDb ID against the Open Movie Database (OMDb) API.\footnote{\url{https://omdbapi.com/}, accessed 1/5/23.} \autoref{tab:data-quantity} shows training data quantities and portions of annotated samples per context while \autoref{tab:data-sample} shows an example of the extracted data. We select six metadata types that we hypothesise to convey useful extra-textual information: \textit{plot description} (which may contain useful topical information), \textit{genre} (which can have an impact on the language used), \textit{year of release} (to account for the temporal dimension of language), \textit{country of release} (to account for regional differences in expression of English), \textit{writers} (to consider writers' style), \textit{PG rating} (which may be associated with e.g. the use of adult language). For validation and testing, we randomly sample $10$K sentence pairs each from the corpus, based on held-out IMDb IDs.

\paragraph{Preprocessing}
The corpus is first detokenised and has punctuation normalised (using Moses scripts \citep{koehn-etal-2007-moses}). Then a custom cleaning script is applied, which removes trailing dashes, unmatched brackets and quotation marks, and fixes common OCR spelling errors. Finally, we perform subword tokenisation via the BPE algorithm with Sentencepiece \citep{kudo-richardson-2018-sentencepiece}.

Film metadata (which comes from OMDb) is left intact except when the fields contain non-values such as \enquote{N/A}, \enquote{Not rated}, or if a particular field is not sufficiently descriptive (e.g. a PG rating field represented as a single letter \enquote{R}), in which case we enrich it with a disambiguating prefix (e.g. \enquote{R} $\rightarrow$ \enquote{PG rating: R}). Regardless of the trained language pair, metadata context is provided in English (which here is either the source or target language). Document-level context is limited to source-side context. Since for *$\rightarrow$\textsc{en} language pairs the context input comes in two languages (e.g. English metadata and French dialogue), we use multilingual models to embed the context in these pairs.

\subsection{Evaluation}
We evaluate the presented approach with the general in-domain test set as well as two external contextual tasks described in this section.

\paragraph{Translation quality} The approaches are evaluated against an in-domain held-out test set of $10$K sentence pairs taken from \textsc{OpenSubtitles}. As metrics, we use \textsc{bleu}\footnote{Computed with SacreBLEU \citep{post-2018-call}.} \citep{papineni-etal-2002-bleu} and \textsc{Comet}\footnote{Computed using the \texttt{wmt20-comet-da} model.} \citep{rei-etal-2020-comet}. 

\paragraph{Control of multiple attributes about dialogue participants (EAMT22)} The EAMT22 task, introduced by \citet{vincent-etal-2022-controlling-extra}, evaluates a model's capability to control information about dialogue participants in English-to-Polish translation. The task requires generating hypotheses that align with four attributes: gender of the speaker and interlocutor(s) (masculine/feminine/mixed), number of interlocutors (one/many), and formality (formal/informal). These attributes can occur in a total of $38$ unique combinations. We investigate whether \textsc{MTCue} can learn this task through zero-shot learning (pre-training on other contexts) or through few-shot learning (when additionally fine-tuned on a constrained number of samples).

To prepare the dataset, we use scripts provided by \citet{vincent-etal-2022-controlling-extra} to annotate \textsc{OpenSubtitles} with the relevant attributes, resulting in a corpus of $5.1$M annotated samples. To leverage the context representation in \textsc{MTCue}, we transcribe the discrete attributes to natural language by creating three sentences that represent the context. For example, if the annotation indicates that the speaker is male, the interlocutor is a mixed-gender group, and the register is formal, we create the following context: (1) \enquote{I am a man}, (2) \enquote{I'm talking to a group of people} and (3) \enquote{Formal}.

We train seven separate instances of \textsc{MTCue} using different artificial data settings. Each setting contains the same number of samples ($5.1$M) but a varying number of \textbf{annotated} samples. To address class imbalances in the dataset (e.g. \textit{masculine speaker} occurring more often than \textit{feminine speaker}) and ensure equal representation of the $38$ attribute combinations, we collect multiples of these combinations. We select sample numbers to achieve roughly equal logarithmic distances: $1$, $5$, $30$, $300$, $3$K and $30$K supervised samples per each of $38$ combinations, yielding exactly $38$, $180$, $1,127$, $10,261$, $81,953$ and $510,683$ samples respectively. Including the zero-shot and full supervision ($5.1$M cases), this results in a total of eight settings. Each model is trained with the same hyperparameters as \textsc{MTCue}, and on the same set of $5.1$M samples, with only the relevant number of samples annotated (non-annotated samples are given as source-target pairs without contexts). We compare our results against our re-implementation of the \textsc{Tagging} approach which achieved the best performance in the original paper \citep[i.e.][]{vincent-etal-2022-controlling-extra}. We train the \textsc{Tagging} model in replicas of the eight settings above.

\paragraph{Zero-shot control of formality (IWSLT22)} We experiment with the generalisation of \mbox{\textsc{MTCue}} to an unseen type of context: formality. In the IWSLT22 formality control task \citep{anastasopoulos-etal-2022-findings}, the model's challenge is to produce hypotheses agreeing with the desired formality (formal/informal). For the English-to-German language pair, the task provides a set of paired examples (each source sentence is paired with a formal reference and an informal one), to a total of $400$ validation and $600$ test examples; for the English-to-Russian pair, only the $600$ test examples are provided. We test the capacity of \mbox{\textsc{MTCue}} to control formality zero-shot, given a textual cue as context input.\footnote{We describe the process of choosing the context input for evaluation in \autoref{app:form}.}

\begin{table*}[h]
\centering
\scalebox{0.6}{
\begin{tabular}{rccccccccccc}
\toprule
Model & \multirow{2}{*}{Params} & \multirow{2}{*}{$d_{model}$} & \multicolumn{3}{c}{Layers} & \multirow{2}{*}{$h$} & \multicolumn{3}{c}{FFN dim.} & \multirow{2}{*}{GPU Hour/Epoch} & \multirow{2}{*}{Epochs to best} \\
& & & Cxt & Src & Dec & & Cxt & Src & Dec \\
 \midrule
\textsc{Base} & $66$M & $512$ & $-$ & $6$ & $6$ & $8$ & $-$ & $2048$ & $2048$ & $-$ & $-$ \\
\textsc{Base-PM} & $107$M & $512$ & $-$ & $10$ & $6$ & $8$ & $-$ & $4096$ & $2048$ & $-$ & $-$  \\ \midrule
\textsc{Tagging} & $107$M & $512$ & $-$ & $10$ & $6$ & $8$ & $-$ & $4096$ & $2048$ & $0.74 \pm 0.35$ & $6.13 \pm 4.09$ \\ 
\textsc{Novotney-cue} & $99$M & $512$ & $6$ & $6$ & $6$ & $8$ & $2048$ & $2048$ & $2048$ & $1.29 \pm 0.56$ & $9.13 \pm 3.60$ \\
\textsc{MTCue} & $105$M & $512$ & $6$ & $6$ & $6$ & $8$ & $2048$ & $2048$ & $2048$ & $0.81 \pm 0.39$ & $9.38 \pm 4.57$ \\ \bottomrule
\end{tabular}}
\caption{Model details for \textsc{MTCue} and baselines. Timings and epochs are averaged across all language directions. }
\label{tab:model-details}
\end{table*}

\subsection{Baselines}
In our experiments, we compare \textsc{MTCue} with three types of baselines:
\begin{enumerate}
    \item \textbf{\textsc{Base} and \textsc{Base-PM}}. These are pre-trained translation models that match \textsc{MTCue} either in the shape of the encoder-decoder architecture  (\textsc{Base}) or in terms of the total number of parameters (\textsc{Base-PM}). For \textsc{Base-PM}, the extra parameters are obtained from enhancing the source encoder, increasing the number of layers ($6\rightarrow10$) and doubling the feed-forward dimension ($2048\rightarrow4096$).
    \item \textbf{\textsc{Tagging}}. Following previous work \citep[e.g.][]{schioppa-etal-2021-controlling,vincent-etal-2022-controlling-extra}, we implement a model that assigns a discrete embedding to each unique context value. Architecturally, the model matches \textsc{Base-PM}. The tags are prepended to feature vectors from the source context and then together fed to the decoder.
    \item \textbf{\textsc{Novotney-cue}}. This baseline is a re-implementation of the \textsc{cue} vectors architecture \citep{novotney-etal-2022-cue} for NMT. It utilises \textsc{DistilBERT} for vectorisation and averages the context feature vectors to obtain the decoder input. In contrast, \textsc{MTCue} employs a parallel attention strategy.
\end{enumerate}

In experiments on formality control, we also report results from the two submissions to the IWSLT22 task, both implementing a supervised and a zero-shot approach:
\begin{enumerate}
    \item \citet{vincent-etal-2022-controlling}. This (winning) submission combines the \textsc{Tagging} approach with formality-aware re-ranking and data augmentation. The authors augment the original formality-labelled training samples by matching sentence pairs from larger corpora against samples of specific formality \citep[akin to the Moore-Lewis algorithm described in][]{moore-lewis-2010-intelligent}. Their zero-shot approach relies on heuristically finding a suitable sample of formality-annotated data similar to the provided set and performing the same algorithm above.
    \item \citet{rippeth-etal-2022-controlling} who fine-tune large pre-trained multilingual MT models with additive control \citep{schioppa-etal-2021-controlling} on data with synthetic formality labels obtained via rule-based parsers and classifiers.
\end{enumerate}
\subsection{Implementation and hyperparameters}
We implement \mbox{\textsc{MTCue}} and all its components in \textsc{Fairseq}, and use HuggingFace \citep{wolf-etal-2020-transformers} for vectorising contexts. We use hyperparameters recommended by \textsc{Fairseq}, plus optimise the learning rate and the batch size in a grid search. We found that a learning rate of $0.0003$ and a batch size of simulated $200$K tokens worked best globally. \autoref{tab:model-details} presents the architecture details and runtimes for the models. All training is done on a single A$100$ $80$GB GPU, one run per model. We use early stopping based on validation loss with a patience of $5$. 

\section{Results} \label{results}
\begin{table*}[ht!]
\resizebox{\linewidth}{!}{
\begin{tabular}{@{}rcccccccccccccccc|cc@{}}
\toprule
\multirow{2}{*}{Model} & \multicolumn{2}{c}{\textsc{en}$\rightarrow$\textsc{de}} 
& \multicolumn{2}{c}{\textsc{en}$\rightarrow$\textsc{fr}} 
& \multicolumn{2}{c}{\textsc{en}$\rightarrow$\textsc{pl}} 
& \multicolumn{2}{c}{\textsc{en}$\rightarrow$\textsc{ru}} 
& \multicolumn{2}{c}{\textsc{de}$\rightarrow$\textsc{en}} 
& \multicolumn{2}{c}{\textsc{fr}$\rightarrow$\textsc{en}} 
& \multicolumn{2}{c}{\textsc{pl}$\rightarrow$\textsc{en}}  
& \multicolumn{2}{c|}{\textsc{ru}$\rightarrow$\textsc{en}} & \multicolumn{2}{c}{Average} \\

 & \textsc{bleu} & \textsc{Comet} & \textsc{bleu} & \textsc{Comet} & \textsc{bleu} & \textsc{Comet} & \textsc{bleu} & \textsc{Comet} & \textsc{bleu} & \textsc{Comet} & \textsc{bleu} & \textsc{Comet} & \textsc{bleu} & \textsc{Comet} & \textsc{bleu} & \textsc{Comet}  & \textsc{bleu} & \textsc{Comet} \\ \midrule
\textit{\textbf{Baselines}}
& & & & & & & & & & & & & & & & & & \\
*\textsc{Base} & $33.60$ & $45.90$ & $34.54$ & $46.92$ & $28.08$ & $58.52$ & $31.37$ & $62.94$ & $39.53$ & $59.56$ & $35.46$ & $55.10$ & $34.42$ & $50.38$ & $39.37$ & $55.99$ & $34.65$ & $54.41$\\
*\textsc{Base-PM} & $34.36$ & $46.77$ & $35.31$ & $48.87$ & $28.66$ & $60.97$ & $32.40$ & $64.55$ & $40.32$ & $60.88$ & $36.16$ & $56.28$ & $35.03$ & $51.77$ & $40.04$ & $\underline{56.86}$ & $35.28$ & $55.87$ \\ \midrule
\textsc{Tagging} & $34.88$ & $49.21$ & $36.74$ & $\underline{51.57}$ & $29.08$ & $\textbf{64.29}$ & $32.32$ & $\underline{65.12}$ & $\textbf{41.52}$ & $\textbf{62.63}$ & $\textbf{37.10}$ & $\textbf{57.41}$ & $\textbf{36.19}$ & $\textbf{53.46}$ & $\textbf{40.33}$ & $\textbf{57.14}$ & $36.02$ & $\textbf{57.60}$ \\
\textsc{Novotney-cue}
& $35.30$ & $49.83$ & $36.75$ & $50.52$ & $29.09$ & $62.69$ & $32.36$ & $\underline{64.90}$ & $40.86$ & $61.91$ & $36.51$ & $56.21$ & $35.28$ & $52.17$ & $39.44$ & $56.08$ & $35.70$ & $56.79$\\ \midrule
\textit{\textbf{Proposed}}
& & & & & & & & & & & & & & \\
\textsc{MTCue} & $\textbf{36.02}$ & $\textbf{50.91}$ & $\textbf{37.54}$ & $\textbf{52.19}$ & $\textbf{29.36}$ & $63.46$ & $\textbf{33.21}$ & $\textbf{65.21}$ & $40.95$ & $61.58$ & $36.57$ & $\underline{56.87}$ & $35.68$ & $52.48$ & $39.97$ & $\underline{56.92}$ & $\textbf{36.16}$ & $57.45$\\ 
\bottomrule
\end{tabular}}

\caption{Translation quality results on the \textsc{OpenSubtitles} test set. *Model trained without access to any context. We highlight the best result in each column and underline all statistically indistinguishable results, $p \leq 0.05$ (except the Average column).}
\label{tab:tq}
\end{table*}
\paragraph{Translation quality}
Results in \autoref{tab:tq} show that \mbox{\textsc{MTCue}} beats all non-contextual baselines in translation quality, achieving an average improvement of $+1.51$ \textsc{bleu}$/$$+3.04$ \textsc{Comet} over \textsc{Base} and $+0.88/$$+1.58$ over \textsc{Base-PM}. It is also significantly better than \textsc{Novotney-cue} ($+0.46/$$+0.66$). \textsc{MTCue} achieves comparable results to the parameter-matched  \textsc{Tagging} model, consistently outperforming it on all language directions from English, and being outperformed by it on directions into English. Since the primary difference between the two models is that \textsc{MTCue} sacrifices more parameters to process context, and \textsc{Tagging} uses these parameters for additional processing of source text, we hypothesise that the difference in scores is due to the extent to which context is a valuable signal for the given language pair: it is less important in translation into English. This is supported by findings from literature: English is a language that does not grammatically mark phenomena such as gender \citep{stahlberg-etal-2007-representation}. 

The largest quality improvements with \textsc{MTCue} are obtained on \textsc{en-de} ($+1.66/$$+4.14$ vs \textsc{Base-PM} and $+1.14/$$+1.70$ vs \textsc{Tagging}) and \textsc{en-fr} ($+2.23/$$+3.32$ vs \textsc{Base-PM} and $+0.80/$$+0.62$ vs \textsc{Tagging}) language pairs. Contrastively, the smallest improvements against \textsc{Base-PM} are obtained on the \textsc{ru-en} pair. \textsc{MTcue} is outperformed by \textsc{Tagging} the most on \textsc{pl-en} ($-0.51/$$-0.98$). As far as training efficiency, \textsc{MTCue} trains significantly faster than \textsc{Novotney-cue}, converging in a similar number of epochs but using significantly less GPU time, on par with \textsc{Tagging} (\autoref{tab:model-details}). Finally, all contextual models considered in this evaluation significantly outperform the parameter-matched translation model (\textsc{Base-PM}), clearly signalling that metadata and document context are an important input in machine translation within this domain, regardless of the chosen approach.

\begin{figure*}[t]
  \centering
  \subfloat[\textsc{MiniLM-v2} embeddings.]
  {
    \includegraphics[width=0.415\textwidth]{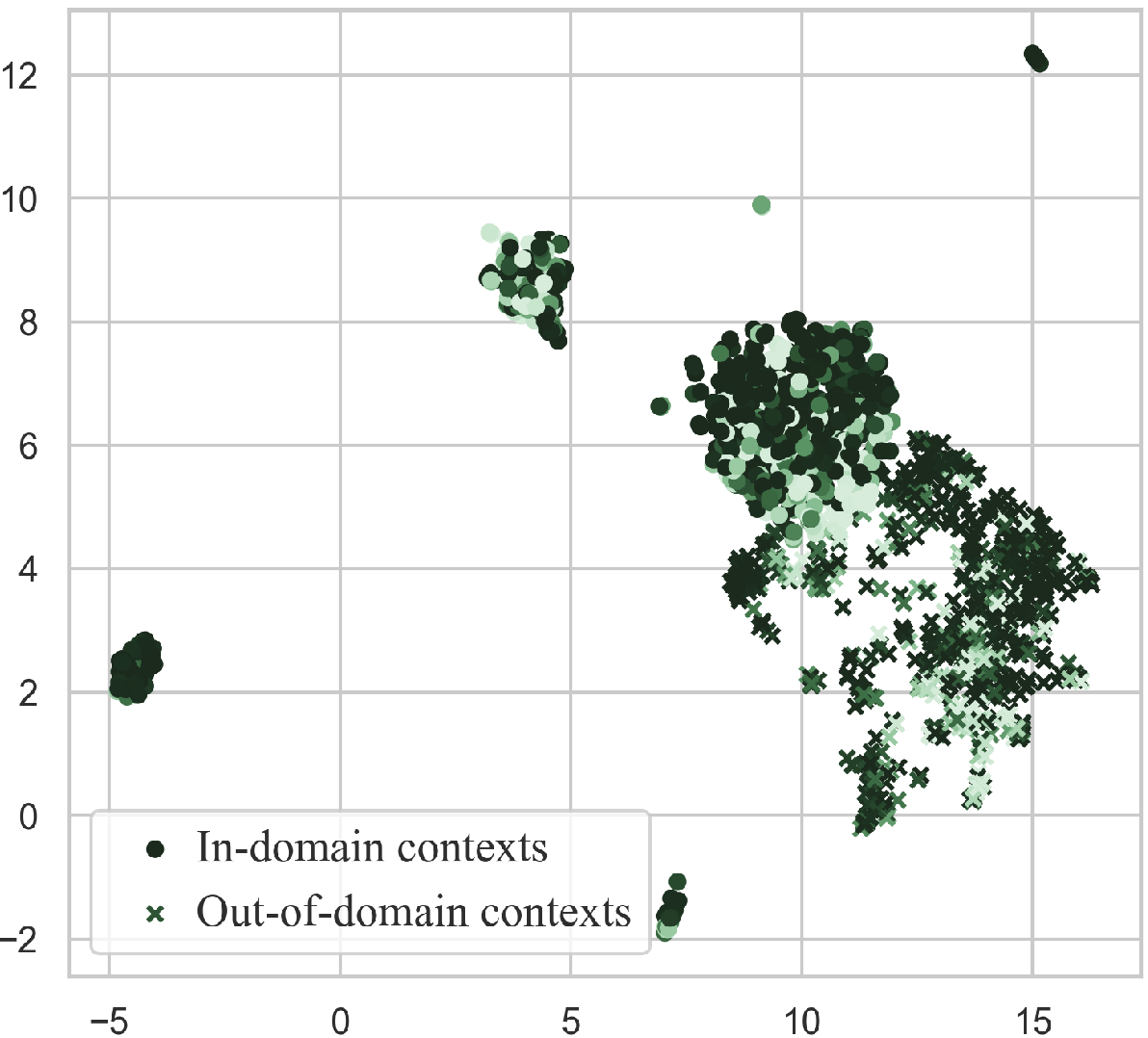}
   \label{fig:form-before}
  }
  \hspace{0.5cm}
  \subfloat[Output of \textsc{MTCue}'s context encoder.]
  {
    \includegraphics[width=0.48\textwidth]{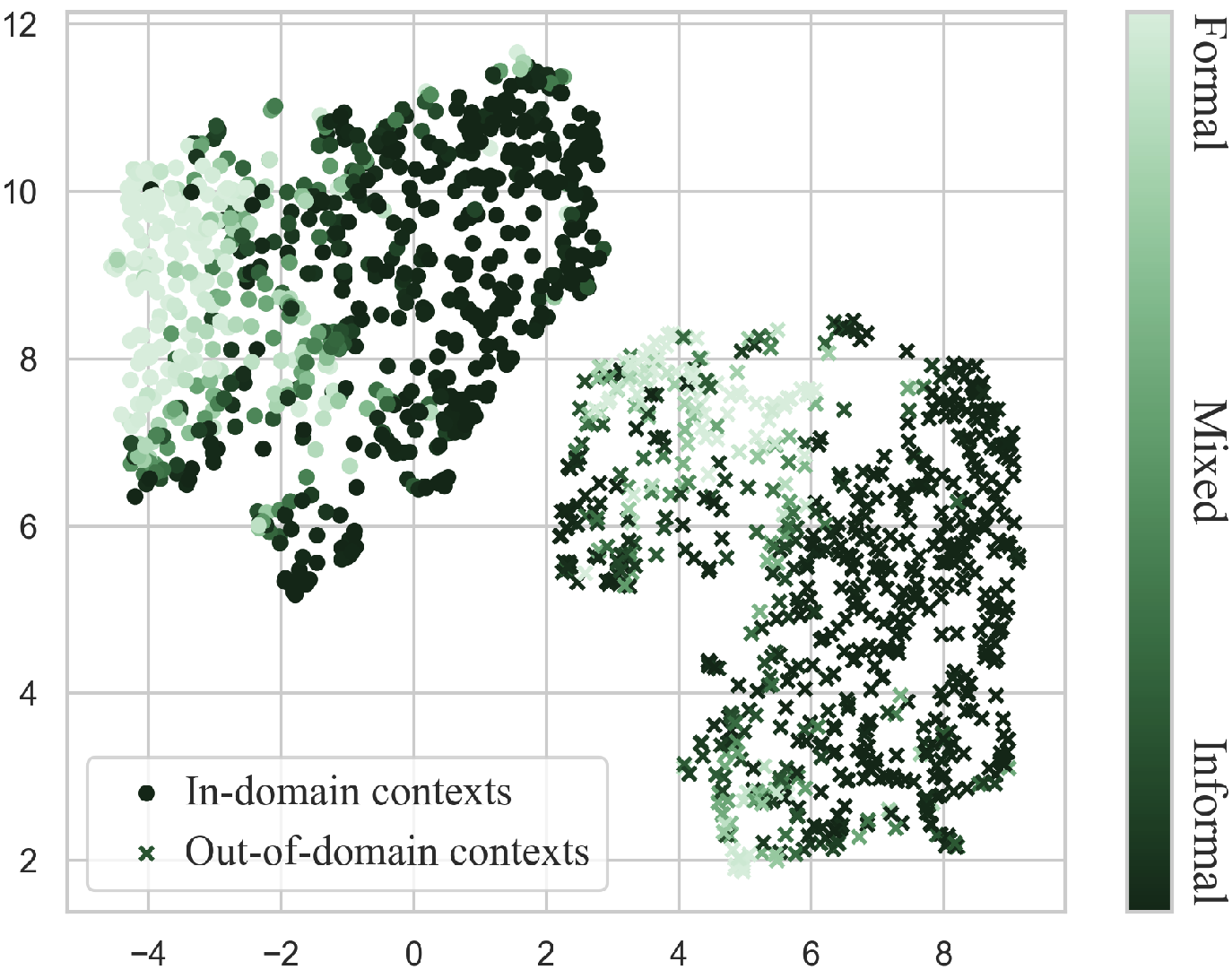}
    \label{fig:form-after}
  }
  \caption{UMAP visualisation of how various contexts impact the formality of produced translations when used as input in \textsc{MTCue}.}
  \label{form-vis}
\end{figure*}

\paragraph{Control of multiple attributes about dialogue participants (EAMT22)}

\begin{figure}[h]
  \centering
  \begin{subfigure}{\columnwidth}
  \centering
    \includegraphics[width=\textwidth]{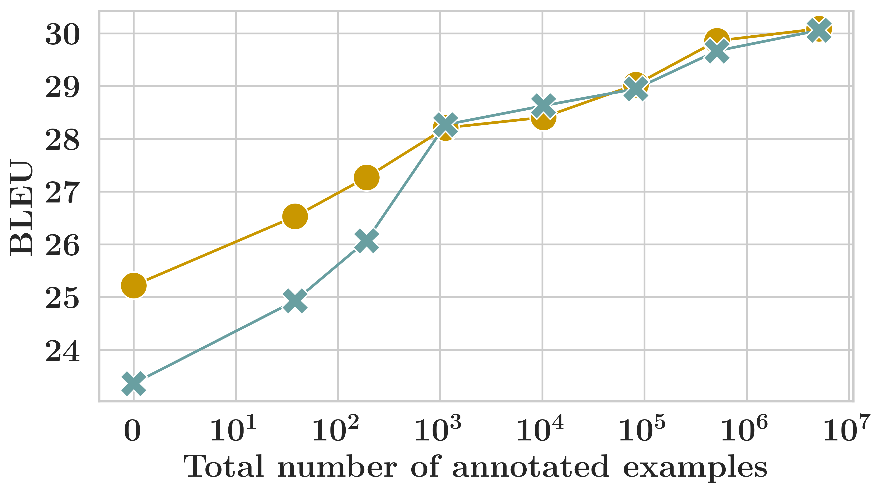}
    \label{fig::eamt-blue}
  \end{subfigure}
   \begin{subfigure}{\columnwidth}
  \centering
    \includegraphics[width=\textwidth]{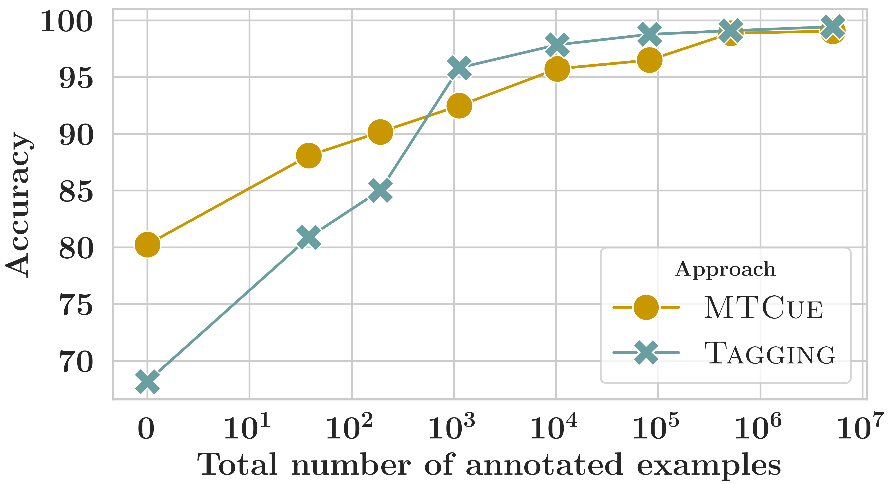}
    \label{fig:eamt-acc}
  \end{subfigure}
  \caption{Evaluation results from the EAMT22 multi-attribute control task.}
  \label{fig:eamt}
\end{figure}

\textsc{MTCue} achieves $80.25$ zero-shot accuracy at correctly translating the speaker and interlocutor attributes, an improvement of $12.08$ over the non-contextual baseline, also expressed in increased translation quality ($25.22$ vs $23.36$ \textsc{bleu}). Furthermore, it bests \textsc{Tagging} at few-shot performance by $5$ to $8$ accuracy points, reaching above $90$\% accuracy with only $190$ of the $5.1$M annotated samples (\autoref{fig:eamt}). Both \textsc{Tagging} and \textsc{MTCue} perform similarly with more supervised data. The \textsc{Tagging} model achieves $+2$ to $+3$ accuracy points in the $1$K to $100$K range, while \textsc{bleu} remains comparable. We hypothesise that this happens because \textsc{MTCue} relies strongly on its pre-training prior when context is scarce: this proves useful with little data, but becomes less relevant as more explicitly labelled samples are added. Finally, with full supervision, both models achieve above $99$\% accuracy.

\paragraph{Zero-shot control of formality (IWSLT22)}

\begin{table}[t]
\centering
\resizebox{\linewidth}{!}{
\begin{tabular}{@{}crlccc@{}}
\toprule
 & Model & Supervision & Formal & Informal & Average \\ \midrule
\multirow{4}{*}{\textsc{en-de}} & Non-context baseline    & $-$ & $74.5$ & $25.5$ & $50.0$ \\
& \citet{rippeth-etal-2022-controlling} & Supervised  & $99.4$ & $96.5$ & $98.0$ \\
& \citet{vincent-etal-2022-controlling} & Supervised  & $100.0$ & $100.0$ & $\textbf{100.0}$ \\ \cmidrule{2-6}
& \textsc{MTCue}                        & Zero-shot   & $100.0$ & $100.0$ & $\textbf{100.0}$ \\ \midrule \midrule
\multirow{4}{*}{\textsc{en-ru}} & Non-context baseline    & $-$ & $96.4$ & $3.6$ & $50.0$  \\
& \citet{rippeth-etal-2022-controlling} & Zero-shot  & $100.0$ & $1.1$ & $50.5$ \\ 
& \citet{vincent-etal-2022-controlling} & Zero-shot  & $99.5$ & $85.8$ & $92.7$ \\ \cmidrule{2-6}
& \textsc{MTCue}                        & Zero-shot   & $100.0$ & $99.4$ & $\textbf{99.7}$ \\ \bottomrule
\end{tabular}}
\caption{Evaluation on the IWSLT22 formality control evaluation campaign. Baseline systems were trained on different corpora.}
\label{tab:form-res}
\end{table}

\textsc{MTCue} appears to successfully control the formality of translations in a zero-shot fashion, achieving nearly $100$\% accuracy on the IWSLT22 test sets across two language pairs, beating all zero-shot models on the \textsc{en-ru} pair and performing on par with the best supervised model for \textsc{en-de}. Notably, both baselines presented in \autoref{tab:form-res} were built to target formality specifically, unlike \textsc{MTCue} which is a general-purpose model.

Following \textsc{MTCue}'s success at controlling formality with sample contexts, we investigate the relationship between context embeddings and their corresponding formality control scores. We consider all $394$ unique contexts from the \textsc{OpenSubtitles} validation data, and another $394$ document contexts (individual past sentences) at random (in-domain). We also use an in-house dataset from a similar domain (dubbing of reality cooking shows with custom annotations of scene contents) and select another $394$ metadata and $394$ document contexts from there (out-of-domain). We run inference on the IWSLT22 test set with each context individually ($1,576$ runs), and use UMAP \citep{umap} to visualise (i) the input embedding from \textsc{MiniLM-v2}, (ii) the output vector of the context encoder and (iii) the corresponding formality score (\autoref{form-vis}).

We invite the reader to pay attention to the separation of dark and light points in \autoref{fig:form-after} that is not present in \autoref{fig:form-before}. There is a spatial property that arises in the context encoder and is shown by \autoref{fig:form-after}, namely a relationship between the feature vectors from context encoder and formality scores across both domains: contexts yielding translations of the same register tend to be clustered together. This is true for both in-domain data (circles) and out-of-domain data (crosses), suggesting that after training this effect generalises to unseen contexts.

For further investigation, we sample a few contexts at random which yield $100$\% zero-shot accuracy (from the \enquote{ends} of the color scale) and find that these contexts tend to have semantic relationships with the type of formality they induce in translations. For example, contexts like \enquote{What's wrong with you?}, \enquote{Wh-what's he doing now?} yield all-informal translations while \enquote{Then why are you still in my office?} or \enquote{I can see you're very interested.} result in all-formal ones. This confirms our hypothesis: \textsc{MTCue}'s context encoder aligns the semantic representation of the input context to the most likely formality it would produce, akin to a human translator deducing such information from available data. Outside of an evaluation scenario like the present one, \textsc{MTCue} may therefore be able to predict from the given context what formality style should be used: an effect only facilitated by the context encoder.

To exemplify how the zero-shot performance of \textsc{MTCue} manifests in practice, we present some examples of outputs for the two tasks in \autoref{app:examples}.

\section{Ablation Study}
\label{ablation}

\begin{table}[ht]
\centering
\resizebox{\linewidth}{!}{
\begin{tabular}{rccccc}
\toprule
\multirow{2}{*}{\textit{Ablation}} &
  \multicolumn{3}{c}{\textsc{Comet}} &
  \multicolumn{2}{c}{\textsc{Zero-shot Accuracy}} \\
 &
  \textsc{en}$\rightarrow$\textsc{de} &
  \textsc{en}$\rightarrow$\textsc{fr} &
  \textsc{en}$\rightarrow$\textsc{pl} &
  IWSLT22 (\textsc{de}) &
  EAMT22 \\
  \midrule
Full \textsc{MTCue}    & $\textbf{46.89}$ & $\textbf{54.06}$ & $62.67$ & $\textbf{100.0}$ & $81.35$ \\
\midrule
no context encoder     & $46.76$ & $53.73$ & $\textbf{63.26}$ & $89.10$ & $77.42$ \\
no pos. embeddings     & $46.68$ & $53.81$ & $62.47$ & $91.65$ & $70.91$ \\
no \textsc{MiniLM-v2}     & $45.32$ & $53.42$ & $62.55$ & $50.00$ & $70.16$ \\
  \midrule
no metadata            & $45.23$ & $53.64$ & $62.64$ & $89.70$ & $\textbf{83.41}$ \\
no doc.-level data     & $46.23$ & $53.49$ & $61.67$ & $68.80$ & $74.64$ \\
random context         & $42.17$ & $51.94$ & $61.74$ & $49.90$ & $68.44$ \\
no context*            & $41.22$ & $50.07$ & $58.94$ & $50.00$ & $67.53$ \\
\bottomrule
\end{tabular}}
\caption{Ablation study on model components and data settings. *Corresponds to non-contextual Transformer.}
\label{tab:abl}
\end{table}

We discuss the robustness of \textsc{MTCue} with an ablation study on the model components as well as a complementary ablation on types of context (metadata vs document). We evaluate three language pairs (\textsc{en}$\rightarrow$\textsc{de,fr,pl}) and report results from single runs (\autoref{tab:abl}): \textsc{Comet} score on the OpenSubtitles18 data and zero-shot accuracy at the two contextual tasks (on the \textbf{validation} sets in all cases).

Removing the context encoder (output of the linear layer is combined with source straight away) or the position embeddings has only a minor effect on the \textsc{Comet} score; replacing $\textsc{MiniLM-v2}$ with a discrete embedding function hurts performance the most. Positional embeddings seem more important to the EAMT22 task than IWSLT22 - possibly because EAMT22 focuses on sentence-level phenomena, so the order of past context matters. 

Replacing \textsc{MiniLM-v2} with a discrete embedding function removes the zero-shot effect in both tasks. An interesting finding is that between metadata and document-level data, it is the latter that brings more improvements to contextual tasks; this means that our model potentially scales to domains without metadata. Finally, using random context degrades performance w.r.t. full model implying that the gains come from signals in data rather than an increase in parameters or training time.

\section{Related Work}
\label{related}

Although contextual adaptation has been discussed in other tasks \citep[e.g.][]{Keskar2019ctrl}, in this section we focus on NMT, as well as set our work side by side with research that inspired our approach.

Existing studies on incorporating context into NMT have primarily focused on document-level context. These approaches include multi-encoder models \citep[e.g.][]{miculicich-etal-2018-document}, cache models \citep{kuang-etal-2018-modeling}, automatic post-editing \citep{voita-etal-2019-context}, shallow fusion with a document-level language model \citep{sugiyama-yoshinaga-2021-context}, data engineering techniques \citep{lupo-etal-2022-divide} or simple concatenation models \citep{tiedemann-scherrer-2017-neural}. Another line of research aims to restrict hypotheses based on certain pre-determined conditions, and this includes formality \citep{Sennrich2016a}, interlocutors' genders \citep[e.g.][]{Vanmassenhove2020, Moryossef2019}, or a combination of both \citep{vincent-etal-2022-controlling-extra}. Other conditions include translation length and monotonicity \citep{lakew-etal-2019-controlling, schioppa-etal-2021-controlling}, vocabulary usage \citep{post-vilar-2018-fast} or domain and genre \citep{matusov-etal-2020-flexible}. While wider contextual adaptation in NMT has been discussed theoretically, most empirical research falls back to gender \citep{Rabinovich2017} or formality control \citep{niu-etal-2017-study}. One exception is \citet{michel-neubig-2018-extreme} who adapt NMT for each of many speakers by adding a \enquote{speaker bias} vector to the decoder outputs.

Our work is motivated by the \textsc{cue} vectors \citep{novotney-etal-2022-cue} and their application in personalised language models for film and TV dialogue \citet{vincent2023personalised}. \textsc{Cue} vectors represent context computed by passing sentence embeddings of the input context through a dedicated encoder. \citeauthor{novotney-etal-2022-cue} show that incorporating \textsc{cue} in language modelling improves perplexity, while \citeauthor{vincent2023personalised} use them to personalise language models for on-screen characters. In contrast, we reformulate \textsc{cue} for contextual machine translation, provide a detailed analysis of incorporating \textsc{cue} into the model, emphasise the importance of vectorising the context prior to embedding it, and examine the benefits for zero-shot and few-shot performance in contextual NMT tasks. 

\section{Conclusions}
\label{conclusions}
We have presented \textsc{MTCue}, a new NMT architecture that enables zero- and few-shot control of contextual variables, leading to superior translation quality compared to strong baselines across multiple language pairs (English to others, cf. \autoref{tab:tq}). We demonstrated that using sentence embedding-based vectorisation functions over discrete embeddings and leveraging a context encoder significantly enhances zero- and few-shot performance on contextual translation tasks. \textsc{MTCue} outperforms the winning submission to the IWSLT22 formality control task for two language pairs, with zero-shot accuracies of $100.0$ and $99.7$ accuracy respectively, without relying on any data or modelling procedures for formality specifically. It also improves by $12.08$ accuracy points over the non-contextual baseline in zero-shot control of interlocutor attributes in translation at the EAMT22 English-to-Polish task. Our ablation study and experiments on formality in English-to-German demonstrated that the context encoder is an integral part of our solution. The context embeddings produced by the context encoder of the trained \textsc{MTCue} can be mapped to specific effects in translation outputs, partially explaining the model's improved translation quality. Our approach emphasises the potential of learning from diverse contexts to achieve desired effects in translation, as evidenced by successful improvements in formality and gender tasks using film metadata and document-level information in the dialogue domain.

\section*{Limitations}
While we carried out our research in four language pairs (in both directions), we recognise that these are mainly European languages and each pair is from or into English. The choice of language pairs was limited by the data and evaluation tools we had access to, however as our methods are language-independent, this research could be expanded to other pairs in the future. 

Another limitation is that the work was conducted in one domain (TV subtitles) and it remains for future work to investigate whether similar benefits can be achieved in other domains, though the findings within language modelling with \textsc{cue} in \citet{novotney-etal-2022-cue} who used a different domain suggest so.

\section*{Ethics Statement}
We do not foresee a direct use of our work in an unethical setting. However, as with all research using or relying on LMs, our work is also prone to the same unwanted biases that these models already contain (e.g. social biases). Therefore, when controlling contextual attributes, researchers should be aware of the biases in their data in order to understand the models' behaviour.

\section*{Acknowledgements}
This work was supported by the Centre for Doctoral Training in Speech and Language Technologies (SLT) and their Applications funded by UK Research and Innovation [grant number EP/S023062/1]. We acknowledge IT Services at The University of Sheffield for the
provision of the High Performance Computing Service. This work was also supported by ZOO Digital.

\bibliography{anthology,mendeley,custom}
\bibliographystyle{acl_natbib}

\appendix

\section{Data Preprocessing}
\label{app:data-prep}
\paragraph{Parsing \textsc{OpenSubtitles}} 
To prepare \textsc{OpenSubtitles} (specifically the document-level part of the corpus), we follow the setup described in \citet{voita-etal-2019-good}. There are timestamps and overlap values for each source-target sample in the corpus; we only take into account pairs with overlap $>= 0.9$ and we use two criteria to build any continuous document: (1) no omitted pairs (due to poor overlap) and (2) no distance greater than seven seconds between any two consecutive pairs. To generate train/validation/test splits, we use generated lists of held-out IMDB IDs based on various published test sets \citep{muller-etal-2018-large, lopes-etal-2020-document, vincent-etal-2022-controlling-extra} to promote reproducibility. These lists can be found within the GitHub repository associated with this paper.

\paragraph{Embedding contexts}
Since a lot of metadata is repeated, and models are trained for multiple epochs, we opt for the most efficient way of embedding and storing data which is to use a memory-mapped binary file with embeddings for unique contexts, and an index which maps each sample to its embedding. This saves more than $90$\% space w.r.t. storing a matrix of all embeddings, and trains over $3 \times$ faster than embedding batches on-the-fly.

\section{Model details}
\label{app:translation}
\textsc{MTCue} is trained from a pre-trained machine translation model (corresponding to the \textsc{Base} model) which is the \texttt{transformer} NMT architecture within \textsc{Fairseq}. We follow model specifications and training recommendations set out by \textsc{Fairseq} in their examples for training a translation model\footnote{\url{https://github.com/facebookresearch/fairseq/tree/main/examples/translation\#iwslt14-german-to-english-transformer}, accessed 1/5/23.}. We train a model for each of the eight language directions on the source-target pairs from \textsc{OpenSubtitles}. We train the model until a patience parameter of $5$ is exhausted on the validation loss. 

\section{Observations on training and hyperparameters} We shortly describe here our findings from seeking the optimal architecture for \textsc{MTCue} and training settings in the hope that this helps save the time of researchers expanding on our work.
\begin{itemize}[noitemsep,leftmargin=*]
    \item Reducing the number of context encoder layers led to inferior performance.
    \item Freezing the source encoder when fine-tuning \textsc{MTCue} from a translation model led to inferior performance,
    \item Training \textsc{MTCue} from scratch $-$ significantly increased training time while having a minor effect on performance.
    \item Other context combination strategies \citep[sequential and flat attention in][]{libovicky-etal-2018-input} led to similar results.
    \item Some alternatives to \textsc{QK-Norm} to combat the problem of the exploding dot-product were successful but had a negative impact on performance:
    \begin{itemize}
        \item using layer normalisation after the linear layer is applied to vectorised contexts,
        \item using SmallEmb\footnote{\url{https://github.com/BlinkDL/SmallInitEmb}, accessed 1/5/23.} which initialises the embedding layer (in our case, the linear proj. layer) to tiny numbers and adds layer normalisation on top.
    \end{itemize}
    \item Zero-shot performance at the IWSLT22 task is generally consistent (at around $98.0-100.0$ accuracy) though may vary depending on the selected checkpoint. We found that training \textsc{MTCue} for longer (i.e. more than $20$ epochs) may improve translation quality but degrade the performance on e.g. this task.
    \item We found that \textsc{MTCue} is generally robust to some hyperparameter manipulation on the \textsc{OpenSubtitles} dataset, and recommend performing a hyperparameter search when training the model on new data. For simplicity, in this paper we use a single set of hyperparameters for all language directions, though for some pairs the results may improve by manipulating parameters such as batch size and context dropout.
\end{itemize}

\label{sec:appendix}

\section{Formality}
\label{app:form}
To evaluate the performance of any tested model on the formality task we had to come up with a fair method of choosing a context to condition on, since in a zero-shot setting the model organically learns the tested attributes from various contexts rather than specific cherry-picked sentences.

To do so, we sampled some metadata from the validation set of the \textsc{OpenSubtitles} data and picked eight contexts (four for the \textit{formal} case and four for the \textit{informal} case) which either used formal or informal language themselves or represented a domain where such language would be used. We also added two generic prompts: \textit{Formal conversation} and \textit{Informal chit-chat}. The full list of prompts was as follows:
\begin{itemize}[noitemsep,leftmargin=*]
    \item Formal:
    \begin{enumerate}
        \item \textit{Formal conversation}
        \item \textit{Hannah Larsen, meet Sonia Jimenez. One of my favourite nurses.}
        \item \textit{In case anything goes down we need all the manpower alert, not comfortably numb.}
        \item \textit{Biography, Drama,}
        \item \textit{A musician travels a great distance to return an instrument to his elderly teacher}
    \end{enumerate}
    \item Informal:
    \begin{enumerate}
        \item \textit{Informal chit-chat}
        \item \textit{I'm gay for Jamie.}
        \item \textit{What else can a pathetic loser do?}
        \item \textit{Drama, Family, Romance}
        \item \textit{Animation, Adventure, Comedy}
    \end{enumerate}
\end{itemize}

We then ran the evaluation as normal with each context separately, and selected the highest returned score for each attribute.

\begin{table*}[ht]
\centering
\resizebox{\textwidth}{!}{
\begin{tabular}{rl}
\toprule
\multicolumn{2}{c}{EAMT22} \\
\midrule
Source         & I just didn't want you to think you had to marry me.            \\
 Context        & \textit{I am a woman. I am talking to a man}           \\
 Reference      & Bo nie chciał\olive{\underline{am}}, żebyś myśl\olive{\underline{ał}}, że cię zmuszam do ślubu. \\
 & \textit{\enquote{Because I didn't want\textsubscript{feminine} you to think\textsubscript{masculine} I am forcing you into a wedding.}} \\
 Baseline       & Po prostu nie chciał\brick{\underline{em}}, żebyś myśl\brick{\underline{ała}}, że musisz \brick{\underline{za mnie wyjść}}. \\
 & \textit{\enquote{I just didn't want\textsubscript{masculine} you to think\textsubscript{feminine} you had to marry\textsubscript{feminine} me.}} \\
 \textsc{MTCue} & Nie chciał\olive{\underline{am}}, żebyś myśl\olive{\underline{ał}}, że musisz \olive{\underline{się ze mną ożenić}}. \\
 & \textit{\enquote{I didn't want\textsubscript{feminine} you to think\textsubscript{masculine} you had to marry\textsubscript{masculine} me.}} \\
                         \midrule
 Source         & So then you confronted Derek. \\
 Context        & \textit{I am talking to a woman} \\
 Reference      & A więc doprowadził\olive{\underline{aś}} do konfrontacji z Derekiem.     \\
 & \textit{\enquote{So then you led\textsubscript{feminine} to a confrontation with Derek.}} \\
 Baseline       & Więc wtedy skonfrontował\brick{\underline{eś}} się z Derekiem. \\
 & \textit{\enquote{So then you confronted\textsubscript{masculine} Derek.}} \\
 \textsc{MTCue} & Więc skonfrontował\olive{\underline{aś}} się z Derekiem. \\
 & \textit{\enquote{So then you confronted\textsubscript{feminine} Derek.}} \\
                         \midrule \midrule
                         \multicolumn{2}{c}{IWSLT22} \\
                         \midrule
 Source         & I got a hundred colours in your city.                           \\
 \textsc{MTCue} (formal) & Ich habe 100 Farben in \underline{Ihrer} Stadt.                    \\
 \textsc{MTCue} (informal) & Ich hab 100 Farben in \underline{deiner} Stadt.                  \\

                         \bottomrule
\end{tabular}}
\caption{Examples of \textsc{MTCue}'s outputs (zero-shot) versus a non-contextual Transformer baseline.}
\label{tab:example-outs}
\end{table*}

\section{Examples of Model Outputs (Zero-Shot)}
\label{app:examples}
We include examples of translations produced zero-shot by \textsc{MTCue} in \autoref{tab:example-outs}. We would like to draw attention particularly to the top example for the EAMT22 task (\enquote{I just didn't want you to think you had to marry me}). The phrase \textit{to marry someone} can be translated to Polish in several ways, indicating that the addressee is to be a wife (\textit{ożenić się z kimś}), a husband (\textit{wyjść za kogoś [za mąż]}) or neutral (\textit{wziąć ślub}). While the reference in this example uses a neutral version, both the baseline model and \textsc{MTCue} opted for feminine/masculine variants. However, the gender of the speaker is feminine, so the phrase \enquote{\textit{... had to marry me}} should use either the neutral version (\textit{wziąć ślub}) or the feminine one (\textit{ożenić się}). The baseline model incorrectly picks the masculine version while \textsc{MTCue} is able to pick the correct one based on the context given. \textsc{MTCue} also correctly translates the gender of the interlocutor: both in the top example (\textit{myślał} vs \textit{myślała}) and the bottom one (\textit{aś} vs \textit{eś}, even though a synonymous expression is used in translation, agreement remains correct). Finally, the IWSLT22 example shows how \textsc{MTCue} produces correct possessive adjectives for each formality. 

\end{document}